\documentclass{article}

     \usepackage[preprint,nonatbib]{neurips_2021}

\usepackage[utf8]{inputenc} 
\usepackage[T1]{fontenc}    
\usepackage{hyperref}       
\usepackage{url}            
\usepackage{booktabs}       
\usepackage{amsfonts}       
\usepackage{nicefrac}       
\usepackage{microtype}      
\usepackage{xcolor}         

\usepackage{graphicx} 
\usepackage{multirow}
\usepackage{amsmath}
\usepackage{amssymb}
\usepackage{bbm}

\newcommand{\etal}{\emph{et al.}}
\newcommand{\eg}{\emph{e.g.}}
\newcommand{\ie}{\emph{i.e.}}
\newcommand{\ProjectName}{MLKD}

\title{Multi-level Knowledge Distillation via Knowledge Alignment and Correlation}

\author{%
  Fei Ding, Yin Yang\\
  Clemson University\\
  \texttt{\{feid, yin5\}@clemson.edu} \\
  \And
  Hongxin Hu \\
  University at Buffalo \\
  \texttt{hongxinh@buffalo.edu} \\
  \AND
  Venkat Krovi, Feng Luo \\
  Clemson University \\
  \texttt{\{vkrovi, luofeng\}@clemson.edu} \\
}

\begin{document}

\maketitle

\begin{abstract}

Knowledge distillation (KD) has become an important technique for model compression and knowledge transfer. In this work, we first perform a comprehensive analysis of the knowledge transferred by different KD methods. We demonstrate that traditional KD methods, which minimize the KL divergence of softmax outputs between networks, are related to the knowledge alignment of an individual sample only. Meanwhile, recent contrastive learning-based KD methods mainly transfer relational knowledge between different samples, namely, knowledge correlation. While it is important to transfer the full knowledge from teacher to student,  we introduce the Multi-level Knowledge Distillation (MLKD) by effectively considering both knowledge alignment and correlation. MLKD is task-agnostic and model-agnostic, and can easily transfer knowledge from supervised or self-supervised pretrained teachers. We show that MLKD can improve the reliability and transferability of learned representations. Experiments demonstrate that MLKD outperforms other state-of-the-art methods on a large number of experimental settings including different (a) pretraining strategies (b) network architectures (c) datasets (d) tasks. 



\end{abstract}


\section{Introduction}

Deep neural networks have recently achieved remarkable success in computer vision~\cite{xie2020self} and natural language processing~\cite{brown2020language}, but they require high computation and memory demand, which limits their deployment in practical applications. KD provides a promising solution to build lightweight models by transferring knowledge from high-capacity teachers with additional supervision signals~\cite{bucilua2006model,hinton2015distilling}. To develop an effective distillation method, there are two main questions: (1) what kinds of knowledge are encoded in a teacher network (2) how to transfer the knowledge between networks. 

Existing KD methods focus on either knowledge alignment or knowledge correlation according to whether the transferred knowledge comes from an individual sample or across samples. The original KD minimizes the KL-divergence loss between the probabilistic outputs of teacher and student networks. This objective aims to transfer the dark knowledge~\cite{hinton2015distilling}, \ie, the assignments of relative probabilities to incorrect classes. Our analysis demonstrates that this logit matching solution actually performs knowledge alignment for an individual sample. Recently, CRD~\cite{tian2019contrastive} has been proposed to learn the structural representational knowledge based on the contrastive objective. SEED~\cite{fang2021seed} is another contrastive distillation method to encourage the student to learn from self-supervised pretrained teachers. The contrastive-learning based methods focus on knowledge correlation because they transfer relational knowledge between different samples. 


\begin{figure}[!t]
\vskip 0.2in
  \centering
  \includegraphics[width=0.8\textwidth]{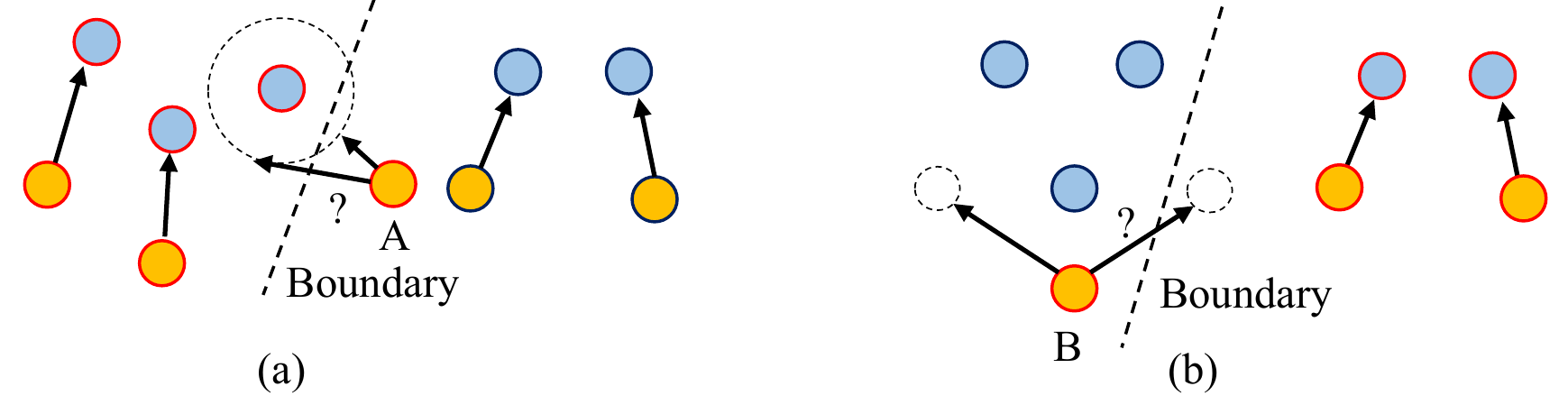}
  \caption{The necessity illustration of knowledge alignment (a) and correlation (b). Blue: teacher, Yellow: student, Red circle: samples with same semantics, Boundary: decision boundary. Knowledge alignment focuses on direct feature matching, and knowledge correlation captures relative relationship between samples. When only one objective is considered, it can result in different mappings (the circle for A and two possible mappings for B), and may not achieve optimal distillation. }
  \label{fig:overview}
\vskip -0.2in
\end{figure}


After dissecting the KD, we find two key factors are neglected in previous works. First, traditional KD relies too much on specific pretraining strategies and network architectures. As new training methods and architectures continue to emerge, we need a more general KD solution. Second, Knowledge alignment and knowledge correlation are largely disconnected in the existing KD methods. We illustrate the necessity of both knowledge alignment and correlation in Figure~\ref{fig:overview}. In this work, we introduce the Multi-level Knowledge Distillation (MLKD) by effectively considering both knowledge alignment and correlation. We propose a feature matching to align the knowledge between the teacher and student. In particular, we find that a spindle-shaped transformation plays a pivotal role in knowledge alignment. We also introduce an effective knowledge correlation solution to capture structural knowledge. Both of them focus on the final feature representation, so our solution (\ProjectName) doesn't depend on specific pretraining tasks or architectures. Besides, we also introduce an optional supervised distillation objective by leveraging the labels, which can be considered to indirectly transfer the category-wise structural knowledge between networks.

The~\ProjectName~enables a student to learn richer representational knowledge from a teacher, which may not be effectively captured by existing methods. We then define a general knowledge quantification metric to measure and evaluate the consistency of visual concepts in the learned representation. Our experiments show that \ProjectName~allows students to learn more generalized representations, certain students can even achieve better performance than teachers. Our method consistently outperforms state-of-the-art methods over a large set of experiments including different pretraining strategies (supervised, self-supervised), network architectures (vgg, ResNets, WideResNets, MobileNets, ShuffleNets), datasets (CIFAR-10/100, STL10, ImageNet, Cityscapes) and tasks (classification, segmentation, self-supervised learning).

\section{Related Work}

\textbf{Knowledge Distillation.}
Hinton~\etal~\cite{hinton2015distilling} first propose KD to transfer dark knowledge from the teacher to the student. The softmax outputs encode richer knowledge than one-hot labels and can provide extra supervisory signals. SRRL~\cite{yang2021knowledge} performs knowledge distillation by leveraging the teacher's projection matrix to train the student's representation via L2 loss. However, these works rely on a supervised pretrained teacher (with logits), and they may be not suitable for self-supervised pretrained teachers. SSKD~\cite{xu2020knowledge} is proposed to combine the self-supervised auxiliary task and KD to transfer richer dark knowledge, but it cannot be trained in an end-to-end training way. Similar to logits matching, intermediate representation~\cite{romero2014fitnets,zagoruyko2016paying,yim2017gift,tung2019similarity,heo2019knowledge} are widely used for KD. FitNet~\cite{romero2014fitnets} proposes to match the whole feature maps, which is difficult and may affect the convergence of the student in some cases. Attention transfer~\cite{zagoruyko2016paying} utilizes spatial attention maps as the supervisory signal. AB~\cite{heo2019knowledge} proposes to learn the activation boundaries of the hidden neurons in the teacher. SP~\cite{tung2019similarity} focuses on transferring the similar (dissimilar) activations between the teacher and student. However, most of these works depend on certain architectures, such as convolutional networks. Since these distillation methods involve knowledge matching in an individual sample, they are related to knowledge alignment. Our work also includes the knowledge alignment objective, and it doesn't rely on pretraining strategies or network architectures.

  

\noindent\textbf{Knowledge distillation and self-supervised learning.} Self-supervised learning~\cite{oord2018representation,bachman2019learning,chen2020simple,he2020momentum,caron2020unsupervised} focuses on learning low-dimensional representations by the instance discrimination, which usually requires a large number of negative samples. Recently, BYOL~\cite{grill2020bootstrap} and DINO~\cite{caron2021emerging} utilize the momentum encoder to avoid collapse without negatives. The momentum encoder can be considered as the mean teacher~\cite{tarvainen2017mean}, which is built dynamically during the student training. For KD, the teacher is pretrained and fixed during distillation. Although different views (augmented images) are passed through networks in self-supervised learning, they are from the same original sample, and have the same semantic meaning. Thus, it performs knowledge alignment between the student and the momentum teacher during each iteration. In particular, DINO focuses on local-to-global knowledge alignment based on multi-crop augmentation.

\noindent\textbf{Relational Knowledge distillation.} Besides knowledge alignment, another research line of KD focuses on transferring relationships between samples. DarkRank~\cite{chen2018darkrank} utilizes cross-sample similarities to transfer knowledge for metric learning tasks.  Also, RKD~\cite{park2019relational} transfers distance-wise and angle-wise relations of different feature representations. Recently, CRD~\cite{tian2019contrastive} is proposed to apply contrastive objective for structural knowledge distillation. However, it randomly draws negative samples and inevitably selects false negatives, hence leading to a suboptimal solution. SEED~\cite{fang2021seed} is proposed to encourage the student to learn representational knowledge from a self-supervised pretrained teacher. But due to the use of a large queue, it cannot effectively transfer knowledge between different semantic samples. Because these distillation methods focus on transferring relational knowledge between different samples, they are related to knowledge correlation. Our work proposes an effective knowledge correlation objective. 


\section{Multi-level Knowledge Distillation}\label{section:MCRD}


For a pair of teacher and student networks, $f_{\eta}^{T}(\cdot)$ and $f_{\theta}^{S}(\cdot)$, the student is trained under extra supervisory signals from the supervised or self-supervised pretrained teacher. $f_{\eta}^{T}(\cdot)$ is the feature extractor and $z_{T}$ denotes the learned last-layer feature. Take the supervised classification task as an example, besides $f_{\eta}^{T}(\cdot)$, there is also a projection matrix $W_{T} \in \mathbb{R}^ {D \times K}$ to map the feature representation to K category logits, where $D$ is the feature dimensionality. We denote by $s(\cdot)$ the softmax function and the standard KD loss~\cite{hinton2015distilling} can be written as:
\begin{align}
\mathcal{L}_{\mathrm{KD}} & = - \sum_{k=1}^{K} s(W_{T}^{k} z_{T}) \log s(W_{S}^{k} z_{S}) \nonumber \\
& = - \sum_{k=1}^{K} s(W_{T}^{k} z_{T}) [\log s(W_{S}^{k} z_{S}) + \log s(W_{T}^{k} h_{\varphi}(z_{S})) - \log s(W_{T}^{k} h_{\varphi}(z_{S}))] \nonumber \\
& = - \sum_{k=1}^{K} s(W_{T}^{k} z_{T}) \log s(W_{T}^{k} h_{\varphi}(z_{S})) + \sum_{k=1}^{K} s(W_{T}^{k} z_{T})  \log \frac{s(W_{T}^{k} h_{\varphi}(z_{S}))}{s(W_{S}^{k} z_{S})},
\end{align}
where $h_{\varphi}(\cdot)$, $z_{S}$ and $W_{S}^{k}$ are trainable, $z_{T}$ and $W_{T}^{k}$ are frozen. $h_{\varphi}(\cdot)$ is a feature transformation function from the student to the teacher. We can observe that when $z_{T} = h_{\varphi}(z_{S})$, the first loss item achieves the optimal solution, and the second loss item becomes the KL divergence between softmax distributions. Therefore, $h_{\varphi}(\cdot)$ plays a pivotal role in minimizing the discrepancy between network's outputs, and simply matching dimensions ~\cite{romero2014fitnets} may not work effectively. First, we prefer to let the student learn excellent features from the teacher, rather than just to minimize the first loss term, so the requirement for $h_{\varphi}(\cdot)$ is that it should not be too powerful. Second, when $h_{\varphi}(\cdot)$ is weak, both of the above loss term becomes large, and make the student more difficult to optimize. Thus, it's crucial to set suitable modeling capability for $h_{\varphi}(\cdot)$.

There are two main limitations to the above objective. First, both loss items depend on the teacher's logits, making this method only suitable for teachers who are pretrained with labels on classification tasks. Thus, it cannot be extended to knowledge transfer from self-supervised pretrained teachers. Second, both loss items focus on feature alignment and minimize the discrepancy between networks' outputs, but ignore important structural knowledge of the teacher. This work proposes to combine knowledge alignment and correlation at the representation level to overcome those two limitations.

\subsection{Knowledge Alignment}

A well-trained teacher already encodes excellent representational knowledge, \ie, categorical knowledge (samples from the same category are close in representation space), the stronger supervision is necessary for better matching between the teacher's representation ($f_{\eta}^{T}(x)$) and the transformation of the student's representation ($h_{\varphi}(f_{\theta}^{S}(x))$). Therefore, we apply the following objective to encourage the student to directly learn the teacher's representation:
\begin{equation}\label{eq:alignment}
\mathcal{L}_{\mathrm{Align}} =\mathbb{E}_{x}\left[\left\|h_{\varphi}(f_{\theta}^{S}(x))-f_{\eta}^{T}(x) \right\|_{2}^{2}\right]. 
\end{equation}
This objective forces the student to directly mimic the teacher's representation, and can provide stronger supervisory signals of inter-class similarities than the standard KD loss~\cite{hinton2015distilling}. Equation~\ref{eq:alignment} focuses only on the matching between the last feature representation. This is different from previous FitNet loss that matches whole feature maps, which will cause training to become difficult or even fail when $h_{\varphi}(\cdot)$ is only regarded as dimensionality matching. In section~\ref{section:exp}, we confirm that appropriate representation capability of $h_{\varphi}(\cdot)$ plays a key role in knowledge alignment.

The knowledge alignment can be further expressed as:
\begin{equation}
\mathcal{L}_{\varphi, \theta} =\mathbb{E}_{x}\left[l \left( h_{\varphi}(f_{\theta}^{S}(x)), g_{\phi}(f_{\eta}^{T}(x))\right)\right], 
\end{equation}
where $l(\cdot,\cdot)$ loss function is used to penalize the difference between networks in different outputs. This is a generalization of existing KD objectives~\cite{hinton2015distilling,romero2014fitnets,yim2017gift,zagoruyko2016paying,yang2021knowledge}. For example, Hinton ~\etal~\cite{hinton2015distilling} calculate KL-divergence between $f^{T}$ and $f^{S}$ in which the linear functions $h_{\varphi}$ and $g_{\phi}$ map representations to logits. SRRL~\cite{yang2021knowledge} utilizes the teacher’s pre-trained projection matrix $W_{T}$ to enforce the teacher’s and student’s feature to produce the same logits via the L2 loss. These methods rely on the logits of the classification task. In contrast, our method is task-agnostic. Although knowledge alignment is effective, it doesn't ensure that the teacher's knowledge is fully transferred, because it only focuses on knowledge transfer for individual sample. 





\subsection{Knowledge Correlation}

The pretrained teacher also encodes rich relationships between samples, and sample relationships transfer allows the student to learn a structure of the representation space similar to the teacher. Here, we propose a knowledge correlation objective to transfer structural knowledge. To be specific, we calculate the relational scores for each (N+1)-tuple samples as the cross-sample relational knowledge. The objective can be expressed as
\begin{equation}
\mathcal{L}_{\mathrm{Corr}}=\sum_{i=1}^{N} l(\psi\left(f_{\eta}^{T}(\tilde{x}_{i}), f_{\eta}^{T}(x_{1}), .., f_{\eta}^{T}(x_{N})\right),
\psi\left(f^{S}(\tilde{x}_{i}), f^{S}(x_{1}), .., f^{S}(x_{N})\right)),
\end{equation}
where N is the batch size, $\psi$ is the relational function that measures the relational scores between the augmented $\tilde{x}_{i}$ and samples $\left\{x_{i}\right\}_{i=1: N}$. $l(\cdot,\cdot)$ is a loss function. The samples in each batch have different semantic similarities, and $\psi$ needs to assign higher scores to samples with similar semantic meaning, assign lower relational scores otherwise. Here, we apply the cosine similarity to measure the semantic similarity between representations, and transform them to softmax distribution for knowledge correlation calculation. All similarities between $\left\{\tilde{x}_{i}\right\}_{i=1: N}$ and $\left\{x_{i}\right\}_{i=1: N}$ can be written as the matrix $\mathcal{A}$. The relational function would be:
\begin{equation}
\psi\left(f(\tilde{x}_{i}), f(x_{1}), .., f(x_{N})\right) = \frac{\exp \left(\mathcal{A}_{i, j} / \tau\right)}{\sum_{j} \exp \left(\mathcal{A}_{i, j} / \tau\right)},   
\end{equation}
where $\tau$ is the temperature parameter to soften peaky distributions and $f(\cdot)$ is the teacher or student network. For the teacher network, $\mathcal{A}_{i, j}$ is calculated by the representations. For the student network, we also apply a transformation function to the representation $\mathbf{z}_{S}$ for loss calculation. We note that this relational function is similar to the InfoNCE loss~\cite{oord2018representation}, which is widely used in self-supervised contrastive learning~\cite{chen2020simple,he2020momentum}. However, our goal is to encode the relationships between samples, not achieve the instance discrimination~\cite{wu2018unsupervised}. Then, we apply the KL-divergence loss to transfer these relationships from the teacher to the student.

In contrast, RKD~\cite{park2019relational} proposes distance-wise and angle-wise losses for relational knowledge distillation. The former has a significant difference in scales and makes training unstable. The latter utilizes a triplet of samples to calculate angular scores (O($N^3$) complexity. Our KL-based solution achieves high-order property with O($N^2$) complexity. SEED~\cite{fang2021seed} is proposed to transfer knowledge from a self-supervised pretrained teacher by leveraging similarity scores between a sample and a queue. However, the large queue results in sparse softmax outputs due to lots of dissimilar samples, which makes it unable to effectively transfer knowledge between different semantic samples. We directly calculate mutual relationships in each batch, and utilize KL divergence loss, which does not require an additional queue and the large-size batch, and has high computation efficiency.

\subsection{Supervised Knowledge Distillation}

Both above objectives are related to feature representations, thus they don't depend on specific pretraining tasks. Here, we also propose an additional distillation objective for supervised pretrained teachers based on the InfoNCE loss. We leverage the true labels to construct positives from the same category and negatives from different categories, which overcomes the sampling bias problem in CRD~\cite{tian2019contrastive}. And there are two kinds of anchors in distillation: teacher and student anchor. The former is from the teacher's outputs, and the corresponding positives and negatives are from the student. The latter is from student's outputs, and its positives and negatives are from the teacher. 
\begin{equation}
\mathcal{L}_{\mathrm{Sup}}^{T/S}=
-\frac{1}{C} \sum_{i=1}^{N} \sum_{j=1}^{2 N} \mathbbm{1}_{i \neq j} \cdot \mathbbm{1}_{\mathbf{y}_{i}=\mathbf{y}_{j}} \cdot \log \frac{\exp \left(\mathbf{z}_{i} \cdot \mathbf{z}_{j} / \tau\right)}{\sum_{k=1}^{2 N} \mathbbm{1}_{i \neq k} \cdot \exp \left(\mathbf{z}_{i} \cdot \mathbf{z}_{k} / \tau\right)}, 
\end{equation}
where $C=2 N_{y_{i}}-1$ and $N_{y_{i}}$ is the number of images with the label $y_i$ in the minibatch. This objective can be considered to indirectly transfer the category-wise structural knowledge between networks. It provides categorical similarities to encourage a student to map samples from the same category into close representation space and samples from different categories be far away. Our formulation is similar to the supervised contrastive loss~\cite{khosla2020supervised}. However, all samples contribute to their gradient calculation while our distillation contains fixed anchors and require further analysis of lower bounds. Besides, the student is also trained with the cross-entropy loss.


\subsection{\ProjectName~objective}

The total distillation loss for any pretraining teacher is a linear combination of knowledge alignment and correlation loss:
\begin{equation}\label{loss:total}
\mathcal{L} = \lambda_{1} \mathcal{L}_{\mathrm{Align}} + \lambda_{2} \mathcal{L}_{\mathrm{Corr}} , 
\end{equation}
where $\lambda_{1}$ and $\lambda_{2}$ are balancing weights. For the supervised pretrained teacher, we also add the above supervised distillation loss $\mathcal{L}_{\mathrm{Sup}}$ and the standard cross-entropy loss $\mathcal{L}_{\mathrm{CE}}$ with balancing weights.



\begin{table*}[t!]
\centering
\caption{Distillation performance comparison between similar architectures. It reports Top-1 accuracy (\%) on CIFAR100 test dataset. We denote the best and the second-best results by \textbf{Bold} and \underline{underline}. The results of all compared methods are from \cite{xu2020knowledge}. }
\label{table:similar}
\begin{tabular}{@{}lcccccc@{}}
\toprule
Teacher & wrn40-2 & wrn40-2 & resnet56 & resnet32$\times$4 &  vgg13 \\
Student & wrn16-2 & wrn40-1 & resnet20 & resnet8$\times$4 & vgg8 \\
\midrule
Teacher  & 76.46 & 76.46 & 73.44 & 79.63  & 75.38 \\
Student  & 73.64 & 72.24 & 69.63 & 72.51  & 70.68 \\
\midrule
KD~\cite{hinton2015distilling}  & 74.92 & 73.54 & 70.66 & 73.33  & 72.98 \\
Fitnets~\cite{romero2014fitnets}  & 75.75 & 74.12 & 71.60 & 74.31  & 73.54 \\
AT~\cite{zagoruyko2016paying} & 75.28 & 74.45& \underline{71.78} & 74.26 & 73.62 \\
FT~\cite{kim2018paraphrasing} & 75.15 & 74.37 & 71.52 & 75.02 & 73.42 \\
SP~\cite{tung2019similarity} & 75.34 & 73.15 & 71.48 & 74.74 & 73.44 \\
VID~\cite{ahn2019variational} & 74.79 & 74.20 & 71.71 & 74.82 & 73.96 \\
RKD~\cite{park2019relational} & 75.40 & 73.87 & 71.48 & 74.47 & 73.72 \\
AB~\cite{heo2019knowledge} & 68.89 & 75.06 & 71.49 & 74.45 & 74.27 \\
CRD~\cite{tian2019contrastive} & \underline{76.04} & 75.52 & 71.68 & 75.90 & 74.06 \\
SSKD~\cite{xu2020knowledge} & \underline{76.04} & \underline{76.13} & 71.49 & \underline{76.20} & \underline{75.33} \\
\midrule
\ProjectName(ours) & \textbf{77.20} & \textbf{76.74} & \textbf{72.34} &\textbf{ 77.11}& \textbf{75.40} \\
\bottomrule
\end{tabular}
\end{table*}

\begin{table*}[th]
\centering
\caption{Distillation performance comparison between different Architectures. It reports Top-1 accuracy (\%) on CIFAR100 test dataset. We denote the best and the second-best results by \textbf{Bold} and \underline{underline}. The results of all compared methods are from \cite{xu2020knowledge}. }
\label{table:cross}
\begin{tabular}{@{}p{2.cm}ccccccc@{}}
\toprule
Teacher & vgg13 & ResNet50 & ResNet50 & resnet32$\times$4 &  resnet32$\times$4 & wrn40-2 \\
Student & MobileV2 & MobileV2 & vgg8 & ShuffleV1 & ShuffleV2 & ShuffleV1 \\
\midrule
Teacher  & 75.38 & 79.10 & 79.10 & 79.63  & 79.63 & 76.46 \\
Student  & 65.79 & 65.79 & 70.68 & 70.77  & 73.12 & 70.77 \\
\midrule
KD~\cite{hinton2015distilling}  & 67.37 & 67.35 & 73.81 & 74.07  & 74.45 & 74.83 \\
Fitnets~\cite{romero2014fitnets}  & 68.58 & 68.54 & 73.84 & 74.82  & 75.11 & 75.55 \\
AT~\cite{zagoruyko2016paying} & 69.34 & 69.28& 73.45 & 74.76 & 75.30 & 75.61 \\
FT~\cite{kim2018paraphrasing} & 69.19 & 69.01 & 73.58 & 74.31 & 74.95 & 75.18 \\
SP~\cite{tung2019similarity} & 66.89 & 68.99 & 73.86 & 73.80 & 75.15 & 75.56 \\
VID~\cite{ahn2019variational} & 66.91 & 68.88 & 73.75 & 74.28 & 75.78 & 75.36 \\
RKD~\cite{park2019relational} & 68.50 & 68.46 & 73.73 & 74.20 & 75.74 & 75.45 \\
AB~\cite{heo2019knowledge} & 68.86 & 69.32 & 74.20 & 76.24 & 75.66 & 76.58 \\
CRD~\cite{tian2019contrastive} & 68.49 & 70.32 & 74.42 & 75.46 & 75.72 & 75.96 \\
SSKD~\cite{xu2020knowledge} & \underline{71.53} & \underline{72.57} & \underline{75.76} & \underline{78.44} & \underline{78.61} & \underline{77.40} \\
\midrule
\ProjectName(ours) & \textbf{72.52} & \textbf{73.18} & \textbf{76.15} & \textbf{78.89} & \textbf{79.54} & \textbf{78.01}\\
\bottomrule
\end{tabular}
\end{table*}

\section{Knowledge Quantification Metric}\label{section:metric}

It's necessary to understand the distilled representation by quantifying the knowledge encoded in networks. Cheng \etal~\cite{cheng2020explaining} proposed to quantify the visual concepts of networks on foreground and background, which requires annotations of the object bounding box. However, these kinds of ground-truth bounding boxes are not always available. Here, we define more general metrics to explain and analyze the knowledge encoded in networks based on the conditional entropy.

Let $\mathbf{X}$ denotes a set of input images. The conditional entropy $H(\mathbf{X}| \mathbf{z} = f(x))$ measures how much information from the input image $x$ to the representation $\mathbf{z}$ is discarded during the forward propagation~\cite{guan2019towards,cheng2020explaining}. A perturbation-based method~\cite{guan2019towards} is proposed to approximate $H(\mathbf{X}|\mathbf{z})$. The perturbed input $\tilde{x}$ follows Gaussian distribution with the assumption of independence between pixels, $\tilde{x} \sim \mathcal{N}\left(x, \mathbf{\Sigma}=\operatorname{diag}\left(\sigma_{1}^{2}, \ldots, \sigma_{n}^{2}\right)\right)$, where $n$ denotes the total number of pixels. Therefore, the image-level conditional entropy $H(\mathbf{X}|\mathbf{z})$ can be decomposed into pixel-level entropy $H_i$ ($H(\mathbf{X}|\mathbf{z})=\sum_{i=1}^{n} H_{i}$), where $H_{i}=\log \sigma_{i}+\frac{1}{2} \log (2 \pi e)$. High pixel-wise entropy $H_i$ indicates that more information is discarded through layers, and the pixels with low pixel-wise entropy are more related with the representation, and the low-entropy pixels can be considered as reliable visual concepts.

We define two general quantification metrics from the view of knowledge quantification and consistency: average and IoU. The average entropy $\bar{H} = \frac{1}{n} \sum_i H_i$ of the image indicates how much information is discarded in the whole input. A smaller $\bar{H}$ indicates that the network utilizes more pixels to compute feature representation from the input. However, more visual concepts don't always lead to the optimal feature representation, which might result in the over-fitting issue~\cite{bengio2013representation}. Ideally, a well-learned network is supposed to encode more robust and reliable knowledge. Thus, we measure the knowledge consistency by the IoU metric, which quantifies the consistency of visual concepts between two views of the same image, \ie, two augmented images $x_1$ and $x_2$.
\begin{equation}
\text{IoU}=\mathbb{E}_{x \in \mathcal{X}}\left[\frac{\sum_{i \in x_1 \cap x_2} \left(S_{\text {concept }}^{\mathrm{1}}(x_i) \cap S_{\text {concept }}^{\mathrm{2}}(x_i)\right)} {\sum_{i \in x_1 \cap x_2} \left(S_{\text {concept }}^{\mathrm{1}}(x_i) \cup S_{\text {concept }}^{\mathrm{2}}(x_i)\right)}\right], \text{where}, S_{\text{concept}}(x)= \mathbbm{1}\left(\bar{H}>H_{i}\right),
\end{equation}
where $\mathbbm{1}$ is the indicator function, and $S_{\text{concept}}(x)$ denotes the set of visual concepts (pixels with lower entropy than $\bar{H}$). $i \in x_1 \cap x_2$ denotes the same pixels of two augmented images. These same pixels are supposed to  obtain similar visual concepts and keep a good consistency between augmented images. Therefore, we choose the ratio between number of visual concepts overlap and number of visual concepts union (IoU) to
measure the knowledge consistency of the learned representations. Our IoU metric meets the requirements of generality and coherency~\cite{cheng2020explaining}, and can be used to quantify and analyze the visual concepts without relying on specific architectures, tasks and datasets.

\section{Experiments}\label{section:exp}


\textbf{Network architectures.} 
We adopt vgg~\cite{simonyan2014very} ResNet~\cite{he2016deep}, WideResNet~\cite{zagoruyko2016wide}, MobileNet~\cite{howard2017mobilenets}, and ShuffleNet~\cite{zhang2018shufflenet} as teacher-student combinations to evaluate the supervised KD on CIFAR100 dataset~\cite{krizhevsky2009learning} and ImageNet dataset~\cite{deng2009imagenet}. Their implementations are from~\cite{tian2019contrastive}. For structured KD, we implement~\ProjectName~based on~\cite{liu2019structured} and evaluate it on Cityscapes dataset~\cite{cordts2016cityscapes}. The teacher model is the PSPNet architecture~\cite{zhao2017pyramid} with a ResNet101 and the student model is set to ResNet18. For self-supervised KD, the teachers are pretrained via MoCo-V2~\cite{chen2020mocov2} or SwAV~\cite{caron2020unsupervised} and we directly download the pretrained weights for our evaluation. The student network is set to smaller ResNet networks (ResNet18, 34). We also perform the transferability evaluation of representations on STL10 dataset~\cite{coates2011analysis} and TinyImageNet dataset~\cite{cs231ntiny, deng2009imagenet}.

\begin{table*}[th]
\centering
\caption{Top-1 and Top-5 error rates (\%) on ImageNet. We denote the best and the second-best results by \textbf{Bold} and \underline{underline}.}
\label{table:imagenet}
\begin{tabular}{@{}l|cc|cccccccc@{}}
\toprule
 & Teacher & Student & SP & KD & AT & CRD & SSKD & SRRL~\cite{yang2021knowledge} & \ProjectName\\
\midrule
Top-1  & 26.70 & 30.25 & 29.38 & 29.34 & 29.30 & 28.83 & 28.38 & \underline{28.27} & \textbf{27.88}\\
Top-5  & 8.58 & 10.93 & 10.20 & 10.12 & 10.00 & 9.87 & \underline{9.33} & 9.40 & \textbf{9.30}\\
\bottomrule
\end{tabular}
\end{table*}

\textbf{Implementation details.} Our implementation is mainly to verify the effectiveness of~\ProjectName. We follow the same training strategy based on existing solutions without any tricks. For supervised KD, we use the SGD optimizer with the momentum of 0.9 and the weight decay of $5 \times 10^{-4}$ in CIFAR100. All the students are trained for 240 epochs with a batch size of 64. The initial learning rate is 0.05 and then divided by 10 at the 150th, 180th and 210th epochs. In ImageNet, we follow the official implementation of PyTorch~\footnote{\url{https://github.com/pytorch/examples/tree/master/imagenet}} and adopt the SGD optimizer with a 0.9 momentum and $1 \times 10^{-4}$ weight decay. The initial learning rate is 0.1 and is decayed by 10 at the 30th, 60th, and 90th epoch in a total of 100 epochs. For these two datasets, we apply normal data augmentation methods, such as rotation with four angles, \ie,  $0^{\circ}, 90^{\circ}, 180^{\circ}, 270^{\circ}$. To perform structured KD, the student is trained with an SGD optimizer with the momentum of 0.9 and the weight decay of $5 \times 10^{-4}$ for 40000 iterations. The training input is set to 512$\times$512, and normal data augmentation methods, such as random scaling and flipping, are used during the training. The self-supervised KD is trained by an SGD optimizer with the momentum of 0.9 and the weight decay of $1 \times 10^{-4}$ for 200 epochs. More detailed training information can be found in the compared methods(CRD~\cite{tian2019contrastive}, SKD~\cite{liu2019structured} and SEED~\cite{fang2021seed}). The temperature $\tau$ in $\mathcal{L}_{\mathrm{Corr}}$ and $\mathcal{L}_{\mathrm{Sup}}$ is set to be 0.5 and 0.07. For the balancing weights, we set $\lambda_1 = 10$ and $\lambda_2 = 20$ according to the magnitude of the loss value. During supervised KD, we set the weights of  $\mathcal{L}_{\mathrm{Sup}}$ and  $\mathcal{L}_{\mathrm{CE}}$ loss to be 0.5 and 1.0. All models are trained using Tesla V100 GPUs on an NVIDIA DGX2 server.

\subsection{Supervised knowledge distillation}

\textbf{CIFAR100.} \ProjectName~is compared with the existing distillation methods ( Table~\ref{table:similar} and Table~\ref{table:cross}). Following CRD~\cite{tian2019contrastive} and  SSKD~\cite{xu2020knowledge}, Table~\ref{table:similar} compares five teacher-student pairs with similar architectures, and Table~\ref{table:cross} focuses on six teacher-student pairs with different architectures. The first two rows represent the classification performance of the teacher and the student being trained individually. For similar-architecture comparisons, \ProjectName~increases the performance of the students by an average of 0.66\% compared to the other best methods. Notably, we find that \ProjectName~enables the student to obtain better performance than the teacher in three out of five pairs. While comparing the teacher-student combinations in different architectures, \ProjectName~still consistently achieves better results than the best competing method (Table~\ref{table:cross}). Our experiments also show that the choice of the student network is also crucial for KD.


\noindent
\textbf{ImageNet.} We conduct one teacher-student pair (teacher: ResNet34, student: ResNet18) on ImageNet. As shown in Table~\ref{table:imagenet}, our \ProjectName~achieves the best classification performances for both Top-1 and Top-5 error rates, which demonstrate the efficiency and scalability on the large-scale dataset.

\subsection{Structured Knowledge Distillation}

Semantic segmentation can be considered as a structured prediction problem, which means that there are different levels of similarities among pixels. To transfer the structured knowledge from the teacher to the student, it's necessary to focus on the pixel-level knowledge alignment and correlation in the feature space. The former encourages the student to learn similar feature representations for each pixel from the teacher, even though their receptive fields (convolutional networks) are different. The latter focuses on maintaining the similarity between pixels belonging to the same class, and the dissimilarity of pixels between different classes. SKD~\cite{liu2019structured} proposes to transfer pair-wise similarities among pixels in the feature space. IFVD~\cite{wang2020intra} proposes to transfer similarities between each pixel and its corresponding class prototype. In contrast, our distillation method can achieve better distillation results than existing structured KD methods (Table~\ref{table:structured}).


\begin{table}[!ht]
\centering
\begin{minipage}[t]{0.45\linewidth}\centering
\caption{The segmentation performance comparison on Cityscapes val dataset. Teacher: ResNet101 and Student:ResNet18.}
\label{table:structured}
\begin{tabular}{lcc}
\toprule
Method & val mIoU (\%) & Params (M) \\
\midrule
Teacher & 78.56 & 70.43 \\
Student & 69.10 & 13.07 \\
SKD~\cite{liu2019structured}  & 72.70 & 13.07 \\
IFVD~\cite{wang2020intra}  & \underline{74.54} & 13.07 \\
\ProjectName (ours) & \textbf{75.73} & 13.07 \\
\bottomrule
\end{tabular}
\end{minipage}\hfill%
\begin{minipage}[t]{0.48\linewidth}\centering
\caption{Top-1 k-NN classification accuracy(\%) on ImageNet. $+$ and $*$indicates the teachers pretrained by MoCo-V2 and SwAV.}
\label{table:self-supervised}
\begin{tabular}{lcc}
\toprule
 Teacher & ResNet18 & ResNet34 \\
\midrule
Supervised  & 69.5 & 72.8 \\
Self-supervised  & 36.7 & 41.5 \\
R-50$^{+}$ + SEED & 43.4 & 45.2 \\
R50x2$^{*}$ + SEED & 55.3 & 58.2\\ 
R50x2$^{*}$ + Ours & \textbf{56.4} & \textbf{59.6} \\ 
\bottomrule
\end{tabular}
\end{minipage}
\end{table}

\subsection{Self-supervised knowledge distillation}

We evaluate the self-supervised distillation with the k-NN nearest neighbor classifier (k=10) as in SEED~\cite{fang2021seed}, which does not require any hyperparameter tuning, nor augmentation. Table~\ref{table:self-supervised} shows the distillation results from different teacher-student pairs. The results of all compared methods are from~\cite{fang2021seed}. The first two rows show the supervised training and self-supervised (MoCo-V2) training baseline results. The k-NN accuracy of self-supervised pretrained ResNet-50(R-50) and ResNet-50w2(R50x2) are 61.9\% and 67.3\%~\cite{caron2021emerging}. We apply the same pretrained R50x2 teacher as~\cite{fang2021seed}, to train students (ResNet18 and ResNet34) using the same training strategy. The results show that our solution can further improve the classification accuracy of students.

\subsection{Ablation Study}


Section~\ref{section:MCRD} demonstrates that it's crucial to set suitable modelling capability for the transformation function $h_{\varphi}(\cdot)$. We apply 2-layer MLPs to implement $h_{\varphi}(\cdot)$ for knowledge alignment and correlation on student's output, which is widely used in self-supervised learning~\cite{chen2020simple,grill2020bootstrap}. We set different dimensions for the hidden layer to model different capabilities in knowledge alignment, which only include $\mathcal{L}_{\mathrm{Align}}$ and $\mathcal{L}_{\mathrm{CE}}$ losses. Table~\ref{table:mlp} shows the comparison results of different multiples of the student representation's dimension (dim($\mathbf{z}_{T}$)). A spindle-shaped MLP (16 times) can achieve the best alignment results. We have not found similar trends in the knowledge correlation, and we directly set all dimensions to dim($\mathbf{z}_{S}$). For the additional $\mathcal{L}_{\mathrm{Sup}}$ and $\mathcal{L}_{\mathrm{CE}}$ losses, only linear projections are used.

\begin{table*}[th]
\centering
\caption{Distillation performance comparison of different $h_{\varphi}(\cdot)$ on the resnet32$\times$4 and ShuffleV2. It reports Top-1 accuracy (\%) on CIFAR100 test dataset. It denotes multiples of dim($\mathbf{z}_{T}$).}
\label{table:mlp}
\begin{tabular}{@{}l|cccccccccc@{}}
\toprule
Hidden size & 0.25 $\times$ & 0.5 $\times$ & 1 $\times$ & 2 $\times$ & 4 $\times$ & 8 $\times$ & 16 $\times$ & 32 $\times$ & 64 $\times$\\
\midrule
Top-1  & 78.54 & 78.63 & 78.58 & 78.62 & 78.43 & 78.57 & \textbf{79.01} & 78.81 & 78.66\\
\bottomrule
\end{tabular}
\end{table*}

\begin{table}[!ht]
\centering
\begin{minipage}[t]{0.45\linewidth}\centering
\caption{Ablation study of \ProjectName. It reports Top-1 accuracy (\%) of two teacher-student pairs on CIFAR100 test dataset.}
\label{table:ablation}
\begin{tabular}{lcc}
\toprule
 Teacher & resnet32$\times$4 & resnet32$\times$4 \\
 Student & resnet8$\times$4 & ShuffleV2 \\
\midrule
$\mathcal{L}_{\mathrm{Align}}$ & 76.59 & 79.01 \\
$\mathcal{L}_{\mathrm{Corr}}$ & 74.94 & 76.06 \\
$\mathcal{L}_{\mathrm{Sup}}$ & 74.73 & 75.98\\
$\mathcal{L}_{\mathrm{Align}} + \mathcal{L}_{\mathrm{Sup}}$ & 76.99 & 79.26 \\
$ \mathcal{L}_{\mathrm{Corr}} + \mathcal{L}_{\mathrm{Sup}}$  & 75.90 & 77.35 \\
$\mathcal{L}_{\mathrm{Align}} + \mathcal{L}_{\mathrm{Corr}}$  & 76.90 & 79.17 \\
All  & \textbf{77.11} & \textbf{79.54} \\
\bottomrule
\end{tabular}
\end{minipage}\hfill%
\begin{minipage}[t]{0.48\linewidth}\centering
\caption{Quantification of knowledge consistency. It reports IoU scores (range from 0 to 1) of two students trained by different distillation methods on CIFAR100 test dataset, and higher is better.}
\label{table:explain}
\begin{tabular}{lcc}
\toprule
 Teacher & resnet32$\times$4 & resnet32$\times$4 \\
 Student & resnet8$\times$4 & ShuffleV2 \\
\midrule
KD  & 0.4647 & 0.2769 \\
CRD  & 0.7288 & 0.4612 \\
$\mathcal{L}_{\mathrm{Align}} + \mathcal{L}_{\mathrm{Corr}} $ &  0.7394 & 0.7449\\
All & \textbf{0.7512}& \textbf{0.7528} \\
\bottomrule
\end{tabular}
\end{minipage}
\end{table}

A student is trained only by the single distillation objective to examine its effectiveness, as shown in Table~\ref{table:ablation}. We find that more objectives can obtain better results, which demonstrates that multiple supervisory signals can improve the representation quality of the student. During three objectives, $\mathcal{L}_{\mathrm{Align}}$ plays a critical role in effective distillation. $\mathcal{L}_{\mathrm{Sup}}$ and $\mathcal{L}_{\mathrm{Corr}}$ can further boost the performance by transferring structural knowledge. Table~\ref{table:explain} compares the knowledge consistency of student networks trained by different distillation methods. It verifies that representation distillation can learn more reliable knowledge, compared with the standard KD method.

\subsection{Transferability of representations}

\begin{table*}[th]
\centering
\caption{Classification accuracy (\%) of STL10 (10 classes) and TinyImageNet (200 classes) using linear evaluation on the representations from CIFAR100 trained networks. We denote compared results from \cite{xu2020knowledge} by *. We denote the best and the second-best results by \textbf{Bold} and \underline{underline}. }
\label{table:linear}
\begin{tabular}{@{}p{1.5cm}ccc|cccc@{}}
\toprule
Dataset & & STL10 & & & TinyImageNet& \\
\midrule
Teacher & resnet32$\times$4 & vgg13 & wrn40-2 & resnet32$\times$4 &  vgg13 & wrn40-2 \\
Student & resnet8$\times$4 & vgg8 & ShuffleV1 & resnet8$\times$4 & vgg8 & ShuffleV1 \\
\midrule
Teacher  & 70.45 & 64.45 & 71.01$^{*}$ & 31.92  & 27.20 & 31.69 \\
Student  & 71.26 & 67.48 & 71.58$^{*}$ & 35.31  & 30.87 & 32.43$^{*}$ \\
\midrule
KD~\cite{hinton2015distilling}  & 71.29 & 67.81 & 73.25$^{*}$ & 33.86  & 30.87 & 32.05$^{*}$ \\
Fitnets~\cite{romero2014fitnets}  & 72.93 & 67.16 & 73.77$^{*}$ & 37.86 & 31.20 & 33.28$^{*}$ \\
AT~\cite{zagoruyko2016paying} & 73.46 & \underline{71.65} & 73.47$^{*}$ & 36.53 & 33.23 & 33.75$^{*}$ \\
FT~\cite{kim2018paraphrasing} & 74.29 & 69.93 & 73.56$^{*}$ & \underline{38.25} & 32.73 & 33.69$^{*}$ \\
SP~\cite{tung2019similarity} & 72.06 & 68.43 & 72.28 & 35.05 & 31.55 & 34.74 & \\
VID~\cite{ahn2019variational} & 73.35 & 67.88 & 72.56 & 37.38 & 31.12 & \underline{35.62} & \\
CRD~\cite{tian2019contrastive} & 73.39 & 69.20 & 74.44$^{*}$ & 37.13 & 33.04 & 34.30$^{*}$ \\
SSKD~\cite{xu2020knowledge} & \underline{74.39} & 71.24 & \underline{74.74}$^{*}$ & 37.83 & \underline{34.87} & 34.54$^{*}$ \\
\midrule
~\ProjectName & \textbf{77.95} & \textbf{74.49} & \textbf{77.43} & \textbf{42.31} & \textbf{38.74} & \textbf{42.48} \\
\bottomrule
\end{tabular}
\end{table*}

We also examine whether the representational knowledge learned by~\ProjectName~can be transferred to the unseen datasets. It performs six comparisons with three teacher-student pairs. The students are fixed to extract feature representations of STL10 and TinyImageNet datasets (all images resized to $32 \times 32$). We then examine the quality of the learned representations by training linear classifiers to perform 10-way and 200-way classification. As shown in Table~\ref{table:linear}, \ProjectName~achieves a significant performance improvement compared to multiple baseline methods, and demonstrates the superior transferability of learned representations. Notably, most distillation methods improve the quality of the student's representations on STL10 and TinyImageNet. The reason why the teacher performs worse on these two datasets may be that the representations learned by the teacher are biased towards the training dataset and are not generalized well. In contrast, \ProjectName~encourages the student to learn more generalized representations.

\section{Conclusion}

In this work, we summarize the existing distillation methods as knowledge alignment and correlation and propose an effective and flexible multi-level distillation method called \ProjectName, which focuses on learning individual and structural representational knowledge. We further demonstrate that our solution can increase the lower bound on  mutual information between distributions of the teacher and student representations. We conduct thorough experiments to demonstrate that our method achieves state-of-the-art distillation performance under different experimental settings. Further analysis of student's representations shows that \ProjectName~can improve the transferability of learned representations. We also demonstrate that our method can work well with limited training data in the few-shot scenario. Due to the hardware limitation, we have not carried out more systematic hyperparameter tuning, which can be done in future works to further obtain better performance. We will ensure that our method is publicly available by maintaining source code online at the GitHub account. Our solution is not related to potential malicious uses, and doesn't have any privacy/security considerations.


%


{
\small
\bibliographystyle{plain}
\bibliography{references}
}

\appendix

\section{Appendix}
\label{appendix}


\subsection{\ProjectName~and mutual information bound}

For a pair of teacher and student networks $f_{\eta}^{T}(\cdot)$ and $f_{\theta}^{S}(\cdot)$, let random variables $T$ and $S$ be the representations of teacher and student ($T=f_{\eta}^{T}(x)$, $S=f_{\theta}^{S}(x)$). We define a distribution $q$ with binary variable $C$ to denote whether a pair of representations $(f_{\eta}^{T}(x_i), f_{\theta}^{S}(x_j))$ is drawn from the joint distribution $p(T, S)$ or the product of marginals $p(T) p(S)$ : $q(T, S | C=1) = p(T, S)$, $q(T, S | C=0) = p(T) p(S)$. The joint distribution indicates positive pairs from close representation space, and the product of marginals indicates negative pairs from far representation space. The previous contrastive representation distillation~\cite{tian2019contrastive} only considers the same input provided to $f_{\eta}^{T}(\cdot)$ and $f_{\theta}^{S}(\cdot)$ as the positives, and samples drawn randomly from the training data as the negatives, which leads to sampling bias problem~\cite{chuang2020debiased}.

Given $N_p$ positive samples and $N_n$ negative samples, we consider the positives in $T$ and $S$ from $p(T, S)$ are empirically related and semantically similar, \eg, representations of the same sample, augmented sample, and samples from the same category, and the negative samples are drawn empirically from different categories. The representation-based KD aims to encourage student's representations to be close to teacher's representations in positives, and those of negatives to be more orthogonal. Then, the priors can be written as: $q(C=1)= N_p/(N_p+N_n)$, $q(C=0)= N_n/(N_p+N_n)$. According to the Bayes' rule, the posterior $q(C=1 | T, S)$ can be written as:
\begin{equation}
q(C=1 | T, S) =\frac{p(T, S)}{p(T, S)+ p(T) p(S) (N_n/N_p)},
\end{equation}
\begin{equation}
\log q(C=1 | T, S) =-\log \left(1 + (N_n/N_p) \frac{p(T) p(S)}{p(T, S)}\right)  
\leq-\log (N_n/N_p)+\log \frac{p(T, S)}{p(T) p(S)}.
\end{equation}

Taking expectation over both sides w.r.t. $q(T, S | C=1)$, we have the mutual information bound as follows:
\begin{equation}\label{eq:mi_bound}
I(T; S) \geq \log (N_n/N_p) + \mathbb{E}_{q(T, S | C=1)} \log q(C=1 | T, S)  
\end{equation}
where $\log (N_n/N_p)$ is a constant term for the given dataset. Previous studies~\cite{tian2019contrastive} suggest that a larger batch size can obtain a better lower bound. But our analysis indicates that the influence factor is the ratio of negative and positive samples, which depends on training data. The second term is to maximize the expectation w.r.t. the student parameters to increase the lower found. But the true distribution $q(C=1 | T, S)$ is intractable. We note that this equation is similar to the InfoNCE loss~\cite{oord2018representation}, which provides a tractable estimator. 

The InfoNCE loss can be written as the softmax formulation. Thus, for knowledge distillation, the loss of each positive pair ($\mathbf{z}_{S}^{i}$, $\mathbf{z}_{T}^{m}$) can be defined as:
\begin{equation}\label{eq:multipe_infonce}
\mathcal{L}_{i, m}=-\log \frac{\exp \left(\mathbf{z}_{S}^{i} \mathbf{z}_{T}^{m} / \tau\right)}{\exp \left(\mathbf{z}_{S}^{i} \mathbf{z}_{T}^{m} / \tau\right)+\sum_{k=1}^{K} \exp \left(\mathbf{z}_{S}^{i} \mathbf{z}_{T}^{k} / \tau\right)},
\end{equation}
where $m$ indicates the $m^{th}$ positive sample $\mathbf{z}_{T}^{m}$ paired with $\mathbf{z}_{S}^{i}$, $k$ indicates the $k^{th}$ negative sample of $\mathbf{z}_{S}^{i}$. Without loss of generality, we set the temperature parameter $\tau=1$ for the following equations. Intuitively, the first term (positive term) encourages the representations of the positives to be similar, while the second term (negative term) encourages representations of the negatives to be more dissimilar. 

When a single positive sample $\mathbf{z}_{T}^{i}$ is paired with $\mathbf{z}_{S}^{i}$ ($M = 1$), and it only relies on positive samples, \ie, same sample or augmented samples, we can obtain the L2-based knowledge alignment objective:
\begin{equation}
\mathcal{L}_{\operatorname{Align}}=-\frac{ \mathbf{z}_{S}^{i} \cdot \mathbf{z}_{T}^{i} }{\left\|\mathbf{z}_{S}^{i}\right\| \cdot\left\|\mathbf{z}_{T}^{i}\right\|} = \frac{1}{2} \cdot \left\|\mathbf{z}_{S}^{i}-\mathbf{z}_{T}^{i}\right\|_{2}^{2} - 1,
\end{equation}
which denotes that we directly maximize the similarity of teacher and student's representations from positives. 

For the knowledge correlation objective, it doesn't directly compare representations between networks. Instead, we consider the relationship between anchor $\tilde{\mathbf{z}}_{T}^{i}$ and one sample $\mathbf{z}_{T}^{j}$ in the teacher by the softmax function:
\begin{equation}
\psi\left(\tilde{\mathbf{z}}_{T}^{i}, \mathbf{z}_{T}^{j}\right) = 
\frac{\exp \left(\tilde{\mathbf{z}}_{T}^{i} \mathbf{z}_{T}^{j} / \tau\right)}{\sum_{k=1}^{N} \exp \left(\tilde{\mathbf{z}}_{T}^{i} \mathbf{z}_{T}^{k} / \tau\right)},
\end{equation}
where $j$ indicates the $j^{th}$ sample. In practice, we calculate the relationships between all samples in the batch. Then we apply KL-based loss to transfer the relationships from the teacher to the student. Because the teacher already encodes the relational knowledge between samples, the KL loss expects to allow the student to learn the similar relationships between samples. Thus it enables the student to map samples from the same category to be closer, and indirectly models the binary classification problem, which is related to $q(C=1 | T, S)$.

Knowledge alignment and correlation objectives don't rely on an explicit definition of positives/negatives, it can be applied in supervised/self-supervised pretrained teachers. In contrast, the supervised knowledge distillation needs true labels to directly identify positives and negatives.

\subsection{The average of conditional entropy}

\begin{table}[t!]
\centering
\caption{Quantification of representational knowledge. It reports average scores of two students trained by different distillation methods on CIFAR100 test dataset.}
\label{table:average}
\begin{tabular}{@{}lccc@{}}
\toprule
 Teacher & resnet32$\times$4 & resnet32$\times$4 & \\
 Student & resnet8$\times$4 & ShuffleV2 & \\
\midrule
KD  & 0.4400 & 0.6307 & \\
CRD  & 0.1460 & 0.4454 & \\
$\mathcal{L}_{\mathrm{Align}}$ & 0.0934 & 0.1641 & \\
$\mathcal{L}_{\mathrm{Corr}}$ & 0.2533 & 0.4288 \\
$\mathcal{L}_{\mathrm{Sup}}$ & 0.2746 & 0.3816\\
All  & \textbf{0.0887} & \textbf{0.1622} & \\
\bottomrule
\end{tabular}
\end{table}

Table~\ref{table:average} shows that the average score $\bar{H}$ of pixel-level conditional entropy as mentioned in Section~\ref{section:metric}. It indicates that the representation of lower $\bar{H}$ tends to achieve better classification performance. A lower $\bar{H}$ also means that the network focuses on more visual concepts to compute the feature representation. In other words, the student can learn richer representational knowledge from the teacher. We utilize the IoU score to quantify the knowledge consistency to evaluate the reliability of visual concepts, as shown in Table~\ref{table:explain}.  Both of average and IoU score can provide more insights about the contrastive representation distillation, in addition to classification evaluation.


\subsection{Linear evaluation of self-supervised KD}

\begin{table*}[th]
\centering
\caption{ImageNet test accuracy(\%) using linear classification. $+$ and $*$indicates the teachers pretrained by MoCo-V2 and SwAV.}
\label{table:self-supervised-linear}
\begin{tabular}{@{}p{2.5cm}c|cc|cc@{}}
\toprule
Methods & & ResNet18 & & ResNet34 & \\
 & Top-1 & Top-1 & Top-5 & Top-1 &  Top-5 \\
 \midrule
Supervised & & 69.5 &  & 72.8 &  \\
Self-supervised  & & 52.5 & 77.0 & 57.4  & 81.6 \\
R-50$^{+}$ + SEED & 67.4 & 57.9 & 82.0 & 58.5  & 82.6 \\
R50x2$^{*}$ + SEED & 77.3 & 63.0 & 84.9 & 65.7 & 86.8 \\
R50x2$^{*}$ + Ours & 77.3 & \textbf{65.8} & \textbf{86.5} & \textbf{67.9} & \textbf{87.7} \\
\bottomrule
\end{tabular}
\end{table*}

We also evaluate the self-supervised KD by linear classification by following previous works in SEED~\cite{fang2021seed}. We apply the SGD optimizer and train the linear classifier for 100 epochs. The weight decay is set to be 0, and the learning rate is 30 at the beginning then reduced to 3 and 0.3 at 60 and 80 epochs. Table~\ref{table:self-supervised-linear} reports the Top-1 and Top-5 accuracy.

\subsection{The visualization of supervised knowledge distillation}

\begin{figure}[t!]
\centering
\includegraphics[width=0.41\textwidth]{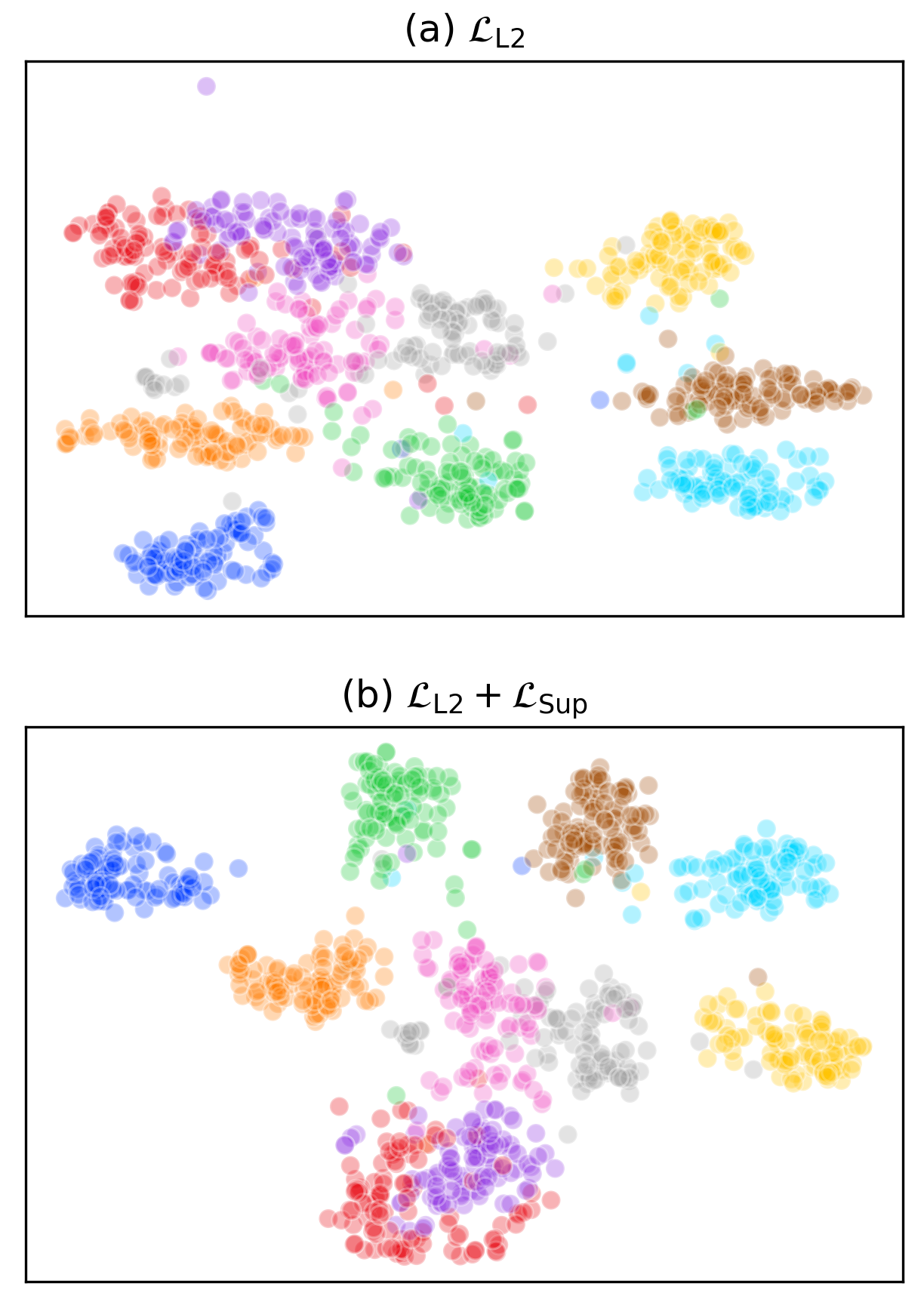} 
\caption{t-SNE visualization of student's representations by applying (a) $\mathcal{L}_{\mathrm{L2}}$ loss and (b) $\mathcal{L}_{\mathrm{L2}} + \mathcal{L}_{\mathrm{Sup}}$ loss. $\mathcal{L}_{\mathrm{Sup}}$ can force the student to map samples of the same category closer in representation space  (teacher: resnet32$\times$4, student: resnet8$\times$4).}
\label{fig:tsne}
\end{figure}

Our paper focuses on multi-level constrastive representation distillation to transfer richer representational knowledge from the teacher to the student. Because the supervised CRD is related to the label information, it's reasonable to verify the categorical knowledge the student have learned by visualization. We randomly select 10 categories from 100 categories for t-SNE visualization, as shown in Figure~\ref{fig:tsne}. After adding $\mathcal{L}_{\mathrm{Sup}}$, it can indeed force students to pull samples from the same category together in the representation space. Thus, the proposed supervised distillation loss enables the student to learn more robust and large margin representations.


\begin{figure}[t!]
\centering
\includegraphics[width=0.44\textwidth]{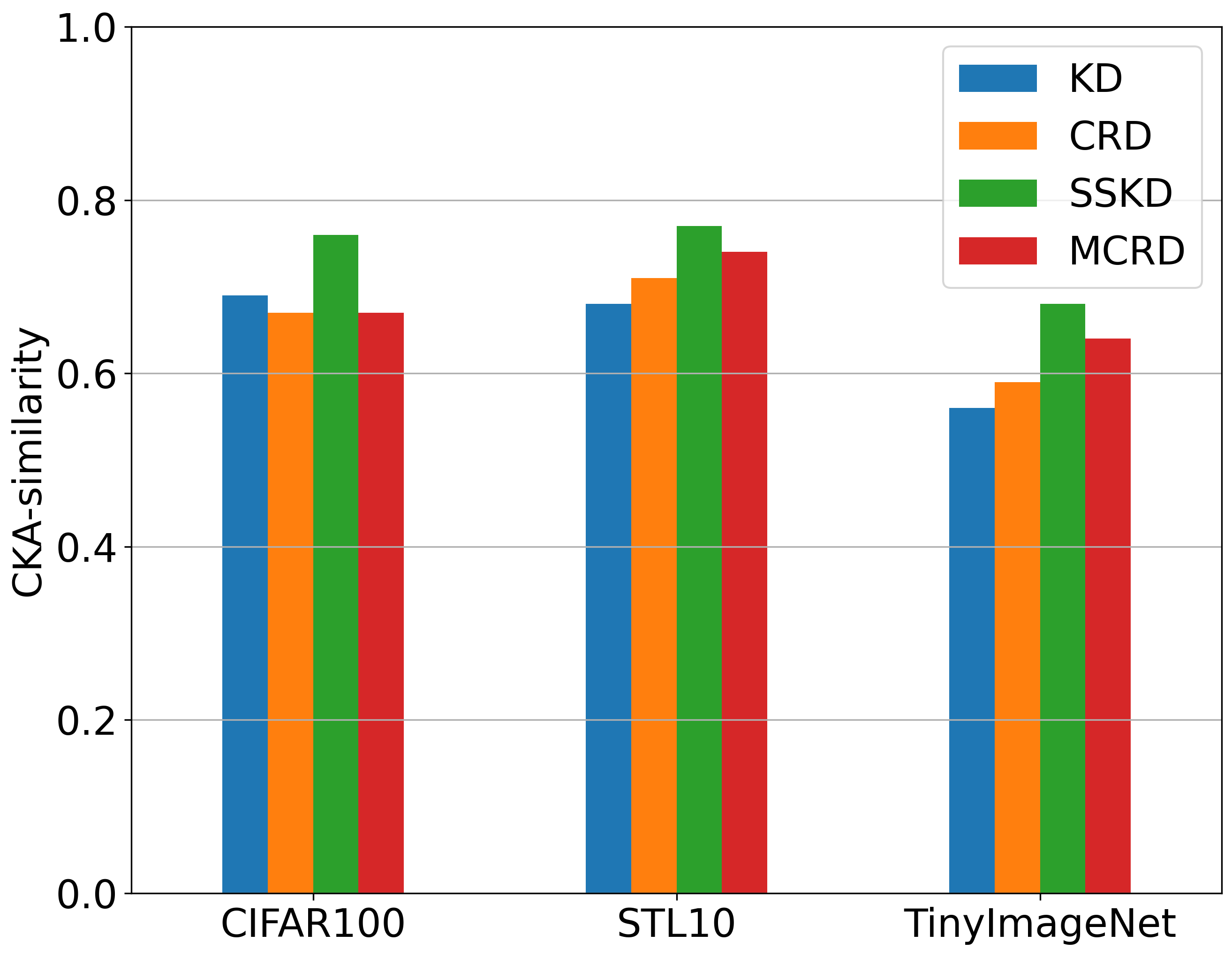} 
\caption{CKA-similarity between the representations from the teacher (vgg13) and student (vgg8) networks.}
\label{fig:similarity}
\end{figure}

\subsection{Teacher-Student similarity}

\ProjectName~can encourage the student to learn richer structured representational knowledge under the multi-level supervisory signals of the teacher. Thus, we conduct the similarity analysis between the teacher's and the student's representations to further understand the contrastive representation distillation. We calculate the CKA-similarity~\cite{pmlr-v97-kornblith19a} (RBF Kernel) between the teacher and student networks, as shown in Figure~\ref{fig:similarity}. Combined with Table~\ref{table:linear}, we find that forcing students to be more similar to teachers does not guarantee that students can learn more general representations.

\subsection{Few-Shot Scenario.} 

\begin{figure}[t!]
\centering
\includegraphics[width=0.42\textwidth]{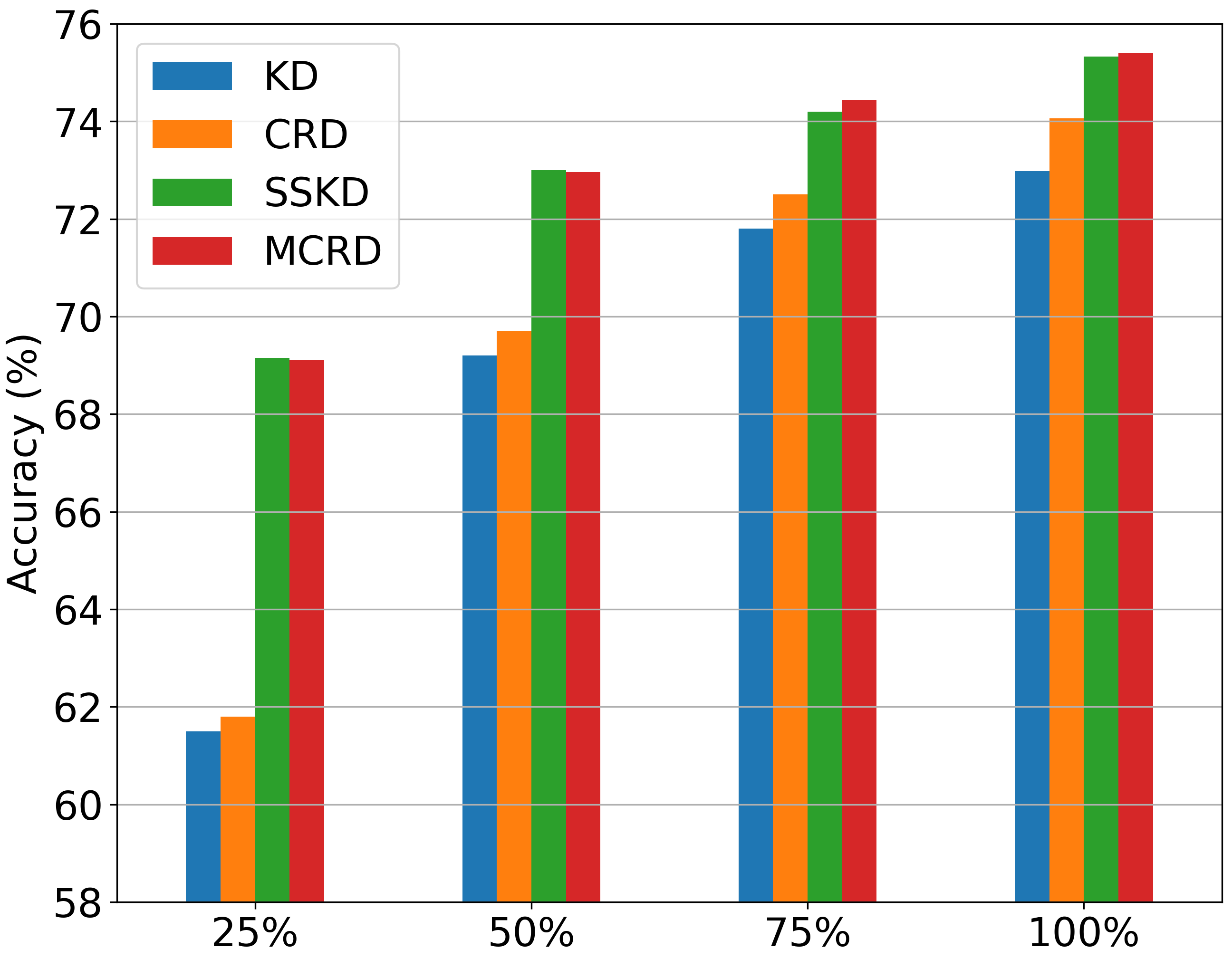} 
\caption{Top-1 accuracy on CIFAR100 test data under a few-shot scenario. The student network is trained with only 25\%, 50\%, 75\% and 100\% of the available training data. }
\label{fig:few-shot}
\end{figure}

\ProjectName~enables the student to learn enough representational knowledge from the teacher, instead of relying entirely on labels. It's necessary to investigate the performance of \ProjectName~under limited training data. We randomly sample 25\%, 50\%, 75\%, and 100\% images from CIFAR100 train set to train the student network and test on the original test set. The comparisons of different methods (Figure~\ref{fig:few-shot}), show that \ProjectName~maintains superior classification performance in all proportions. As the training set size decreases, multi-level supervisory signals in \ProjectName~serve as an effective regularization to prevent overfitting.

\end{document}